\relax
\documentclass{article}
\usepackage{arxiv}
\usepackage[utf8]{inputenc} 
\usepackage[T1]{fontenc}    
\usepackage{booktabs}       
\usepackage{amsfonts}       
\usepackage{nicefrac}       
\usepackage{microtype}      
\usepackage{doi}

\usepackage{times}  
\usepackage{helvet} 
\usepackage{courier}  
\usepackage{graphicx} 
\usepackage{caption} 
\title{Bilevel Continual Learning}
\usepackage{times}

\usepackage{soul}
\usepackage{url}
\usepackage{graphicx}
\usepackage{amsmath}
\usepackage{booktabs}
\usepackage{tabularx}
\usepackage{booktabs}
\usepackage{comment}
\newcolumntype{C}{>{\centering\arraybackslash}X} 
\usepackage{nameref} 
\usepackage{amsmath}
\usepackage[italicdiff]{physics}
\usepackage{algorithm}
\usepackage[noend]{algpseudocode}
\makeatletter
\def\BState{\State\hskip-\ALG@thistlm}
\makeatother

\DeclareMathOperator{\E}{\mathbb{E}}

\usepackage{mathtools}

\usepackage{amsfonts}
\usepackage{booktabs}
\usepackage{siunitx}
\usepackage{color}
\usepackage{amsthm}
\newtheorem{theorem}{Theorem}

\usepackage{amsmath}
\author{Ammar Shaker, Francesco Alesiani, Shujian Yu, Wenzhe Yin\thanks{ Work performed while at NEC Laboratories Europe GmbH}\\ 
NEC Laboratories Europe GmbH\\ 
Kurfuerstenanlage 36\\
D-69115 Heidelberg\\
\{Ammar.Shaker,Francesco.Alesiani\}@neclab.eu, Wenzhe.Yin@stud.uni-heidelberg.de 
}

\begin{document}
\maketitle

\begin{abstract}
Continual learning (CL) studies the problem of learning a sequence of tasks, one at a time, such that the learning of each new task does not lead to the deterioration in performance on the previously seen ones while exploiting previously learned features.
This paper presents Bilevel Continual Learning (BiCL), a general framework for continual learning that fuses bilevel optimization and recent advances in meta-learning for deep neural networks. BiCL is able to train both deep discriminative and generative models under the conservative setting of the online continual learning. Experimental results show that BiCL provides competitive performance in terms of accuracy for the current task while reducing the effect of catastrophic forgetting. \emph{This is a concurrent work with \cite{pham2020bilevel}. We submitted it to AAAI 2020 and IJCAI 2020. Now we put it on the arxiv for record. Different from \cite{pham2020bilevel}, we also consider continual generative model as well. At the same time, the authors are aware of a recent proposal on bilevel optimization based coreset construction for continual learning \cite{borsos2020coresets}.}
\end{abstract}


\section{Introduction}
Human Intelligence (HI) shows the ability to leverage past tasks for solving new ones while retaining the capability of solving previously examined tasks. 
On the course of giving machines the ability to successively learn on sequences of tasks, \emph{catastrophic forgetting}~\cite{mccloskey1989catastrophic,ratcliff1990connectionist} prohibits the learners from accumulating new knowledge without overwriting previously acquired ones.
Continual learning addresses the question of how to retain or even improve the performance on previous tasks while not hindering the learning capability for new tasks (similar to neuroplasticity), without the need of going through old experiences once again.

In continual learning~\cite{thrun:1996:learning,ring_continual_1994}, i.e., learning from a sequence of tasks, \emph{catastrophic forgetting}~\cite{mccloskey1989catastrophic,ratcliff1990connectionist} occurs when learning a new task is likely to override the model parameters (e.g. neural networks' parameters) that have been learned in the past, thus, causing performance degradation on learned past tasks. 
Distillation of information \cite{ferrari_lifelong_2018}
and dreaming \cite{hu_overcoming_2018} are common approaches to extract knowledge and make it available for future and past tasks.  
Regularization-based continual learning methods employ parameters learned from past tasks for penalization terms. For example, \cite{kirkpatrick_overcoming_2016} penalizes the update of parameters based on their relevance to past tasks. Alternatively, continual learning can be implemented directly by extending the optimization problem with cost terms that use a limited memory from past samples. 
For example, in \cite{lopez-paz_gradient_2017,riemer_learning_2018}, samples are used to steer the gradient in a direction of positive transfer learning. 

In this paper, we introduce a novel continual learning framework by re-formulating the continual training of consecutive tasks as a bilevel optimization problem ~\cite{colson2007overview}. 
We show that the proposed framework reduces catastrophic forgetting on previous tasks by requiring gradient updates in directions that are beneficial for new and previous tasks, while allowing training on the current task, thanks to the bilevel nature of the formulation.  
The general formulation shows how discriminative, as well as generative problems, can be handled within our framework. Empirical results show the validity of our claims and the effective reduction in catastrophic forgetting. 

BiCL is a general continual learning framework that fulfills the most important desiderata of ideal continual learning, namely forward and backward transfer under constant memory constraints with minimal catastrophic forgetting. The contribution of this paper is multi-fold:
\begin{itemize}
	\item \textbf{Bilevel CL framework:} We introduce a method to incorporate the continual learning loss into a bilevel problem and propose an algorithm that approximates its solution;
	\item \textbf{Discriminate and Generative CL methods:} We provide theoretical and practical guidance on how our framework can be used for both: the discriminative and generative cases;
    \item \textbf{Accurate Prediction and Retained Accuracy:} The experimental results show that BiCL achieves often superior performance on both the current task and the previous tasks even under the strong constraint of very small memory size.
\end{itemize}

In the following, we first present a short overview of related work and motivate the application of bilevel optimization in machine learning. In Section \nameref{Bilevel_Continual_learning_framework}, we introduce our general framework, show its mathematical formulation, and how it achieves the goal of forward/backward transfer. Section \nameref{Experiments} presents a practical way of implementing the framework followed by the experimental results of the framework. Section \nameref{ Conclusions} concludes the paper.

\section{Related Work}
\label{related_work}
Before introducing our approach, we present an overview of related work concerning two main aspects of the paper: (1) Continual learning and catastrophic forgetting in deep neural networks (DNNs); and (2) Bilevel Optimization in the context of deep learning.

\subsection{Continual Learning and Catastrophic Forgetting in Neural Networks}
Several continual learning approaches have been proposed in recent years addressing the problem of catastrophic forgetting. We briefly review the most relevant ones to our work. Interested readers can refer to~\cite{li2017learning,parisi2019continual} for comprehensive surveys on this topic.

\textbf{Regularization-based approaches} These approaches attempt to identify ``important” parameters for previous tasks and penalize changes to those parameters while learning new tasks. For example, Elastic Weight Consolidation (EWC)~\cite{kirkpatrick_overcoming_2016} uses the Fisher information matrix (FIM) as a measure of parameter importance for previous tasks. The performance of EWC has been significantly improved in~\cite{liu2018rotate} using a network reparameterization trick to obtain a better diagonal approximation of the FIM. In a parallel work, Synaptic Intelligence (SI)~\cite{zenke2017continual} estimates the importance of the network parameters during task learning through accumulating the contribution of each parameter to the change in the loss. Recently,~\cite{chaudhry2018riemannian} suggests a KL-divergence based generalization of EWC and SI, whereas~\cite{serra_overcoming_2018} uses an attention mechanism to activate/deactivate parts of the network for the learning of new tasks.

\textbf{Coreset-based models} These approaches alleviate the constraint on the availability of data by allowing the storage of a few samples from previous data (which are called coreset). For example, iCaRL~\cite{rebuffi_icarl_2017} stores $2,000$ samples from previous batches and rely on a mixture of cross-entropy and distillation loss to alleviate forgetting. Other examples include Gradient Episodic Memory (GEM)~\cite{lopez-paz_gradient_2017} and its extension Averaged GEM (AGEM)~\cite{chaudhry_efficient_2019}: both methods assume that a coreset of past data can be stored and used in the future.

\textbf{Generative models} Inspired by biological mechanisms that the hippocampus encodes and replays recent experiences to help the memory in the neocortex to consolidate~\cite{o2002hippocampal}, a natural approach to overcome catastrophic forgetting is to produce samples of previous data that can be added to the new data to learn a new task. For example, Deep Generative Replay (DGR)~\cite{shin2017continual} trains a generative model on each task's data, and uses it to generate pseudo-data that substitute the memory of previous tasks. A recent trend in this direction is to replay features using feature extractor pre-trained on large-scale data sets and replay features~\cite{hu_overcoming_2018,xiang2019incremental}. In general, those methods present good results but require complex models to be able to generate reliable data. In \cite{beaulieu2020learning}, a "neuromodulatory" network is used to gate the accumulated information; while different from generative approaches, this method requires a secondary network similar to the generative models, thus aggravating the requirement in term of parameters of the network.

\textbf{Meta-learning based approaches} Combining multiple stochastic gradient descendant (SGD) steps has been proposed by Riemer et. al \cite{riemer_learning_2018} for continual learning problems; their approach, Meta-Experience Replay (MER), 
integrates a reply of past experience while learning new tasks. As a result, the optimized objective function is being regularized by a term that forces gradients (on new examples) to have transfer and less interference with past ones.
Another emerging class of methods is continual learning via meta-learning; among these methods is online aware Meta-learning (OML) \cite{javed2019meta} that learns two parameter sets using partial update rule. This method expects tasks to be sampled from a common distribution; thus, the same task might appear multiple times.

Recently, {\bf generative continual learning } has received increasing attention. A naive way would be to directly apply the variational auto-encoders (VAE) model \cite{kingma_auto-encoding_2013} to the new task's data $\mathcal{D}_{t}$ with the parameters initialized at the parameter $\theta_{t-1}$ found for the previous task. 
However, similar to discriminative approaches, continual learning of deep generative models also suffers catastrophic forgetting. Recent work \cite{nguyen_variational_2017} shows that EWC \cite{kirkpatrick_overcoming_2016} can be directly adapted for the VAE model in a continual learning scenario to alleviate the catastrophic forgetting.


\subsection{Bilevel Optimization in the Context of Deep Learning}
\label{Bilevel_Optimization_in}
Bilevel problems raise when the solution for the main variables of a minimization problem is subject to the optimality of a secondary minimization problem, whose variables also depend on the main variables. The two problems are also called outer and inner problems. The idea is that the solution of the outer problem depends on the solution of the inner problem, where the decision is taken by the master (outer level), and the follower (inner level) acts by adopting the value found by the master.

In machine learning, hyper-parameter optimization tries to find the predictive model's parameters $w$, with respect to the vector of hyper-parameters $\lambda$ that minimizes the validation error. This can be mathematically formulated as the bilevel problem \cite{franceschi_bilevel_2018}:
\begin{subequations} \label{eq:biproblem}
	\begin{eqnarray}
	\min _{\lambda} F(\lambda) &=& \E_{s \sim D^{\text{val}}} \{ f(w_{\lambda},\lambda,s)\}  \\
	\text{s.t.} ~ w_{\lambda} &=& \arg \min_{w} \E_{s' \sim D^{\text{tr}}} \{ g(w,\lambda,s') \} \ . \label{biproblem} 
	\end{eqnarray}
\end{subequations}

The outer objective tries to minimize the validation error $E_{s \sim D^{\text{val}}} \{ f(w_{\lambda},\lambda,s) \}$ in the space of the hyper-parameters on validation data $D^{\text{val}}$, whereas $E_{s \sim D^{\text{tr}}} \{ g(w,\lambda,s)\}$ is the regularized empirical error on the training data $D^{\text{tr}}$ and $D^{\text{val}} \bigcup D^{\text{tr}} = D$, see \cite{franceschi_bilevel_2018}. The sub problem is solved using gradient descendant method and its variations. 
The bilevel optimization formulation has the advantage of allowing to optimize two different cost functions (in the inner and outer problems) on different data (training/validation), thus, alleviating the problem of over-fitting and implementing an implicit cross-validation procedure.

In the context of deep learning, bilevel optimization has been adopted by many works to search for the hyper-parameters under the performance constraint on a validation set. Kunisch and Pock \cite{kunisch2013bilevel} apply bilevel optimization to learn the parameters of a variational image denoising model. In \cite{jenni2018deep}, the principle of cross-validation is formulated as a bilevel optimization problem to train deep neural networks for better generalization capability and reduced test errors. More recently, Franceschi et. al \cite{franceschi_bilevel_2018} propose a bilevel optimization framework that unifies gradient-based hyper-parameter optimization and meta-learning. More specifically, the authors reformulate the framework of meta-learning in a way that treats the weights of the output layer of a neural network as variables in the inner problem, which minimizes empirical loss on training sets over multiple tasks. The weights of the remaining hidden layers (seen as hyper-parameters) are treated as variables in the outer problem, which minimizes the validation error on the validation data over multiple tasks.

\section{Bilevel  Continual Learning Framework}
\label{Bilevel_Continual_learning_framework}
We start by introducing the problem of continual learning, and a loss function that incentivizes positive transfer learning and reduces catastrophic forgetting. Thereafter, we formally introduce the bilevel framework on the discriminative and generative setting. Finally, Section \nameref{sec:bilevel-reptile} shows how the framework is extended to bridge the gap and overcome catastrophic forgetting.

\subsection{Continual Learning Problem}
\begin{figure}[h]
	\centering
	\includegraphics[width=6cm]{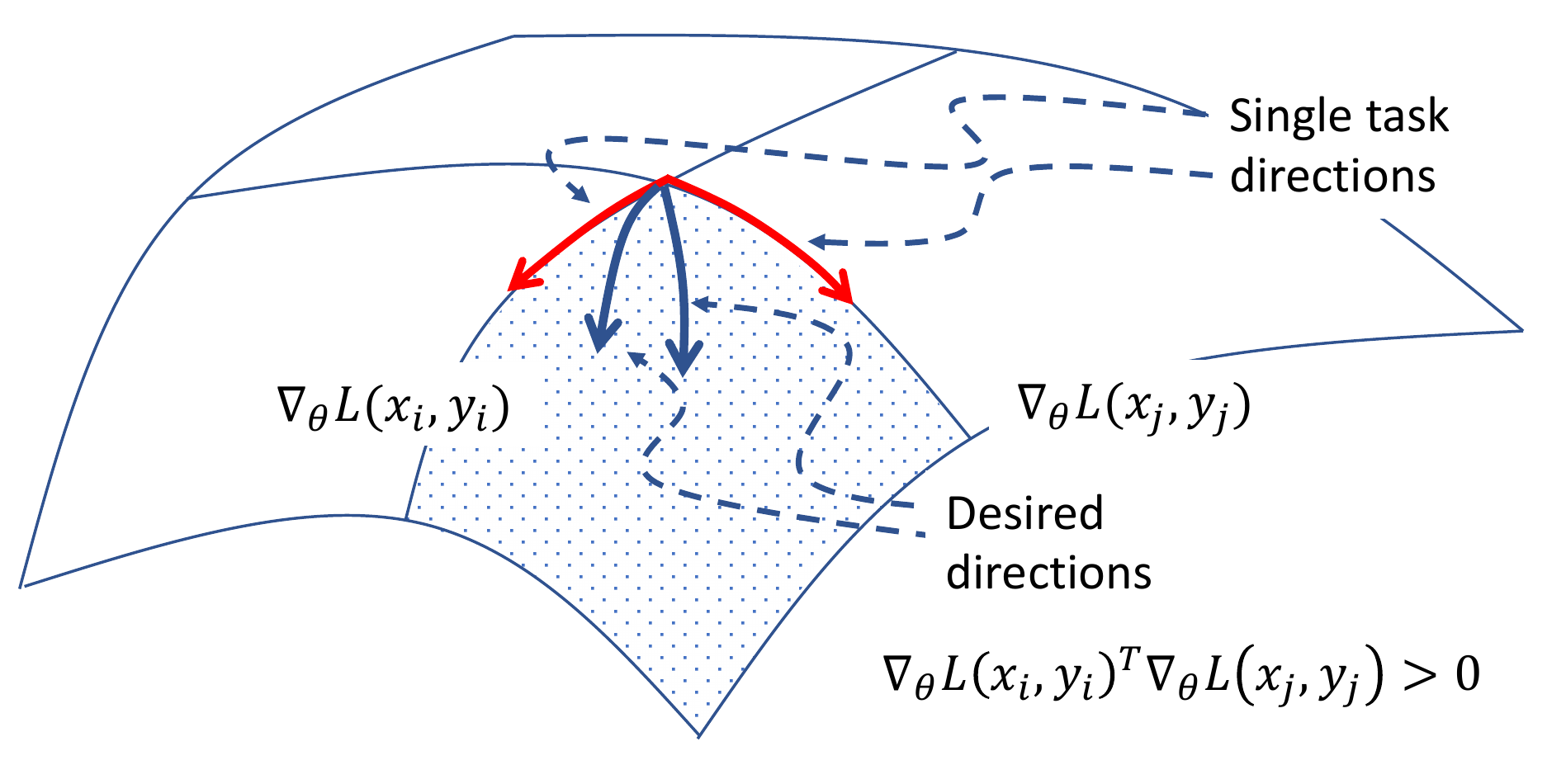}
	\caption{Continual learning loss function and desirable directions. Red lines represent single task gradient directions, blue lines represent multi-task gradient directions}
	\label{fig:cl_loss}
\end{figure}

In continual learning, we are interested in positive transfer learning and avoiding negative transfer among tasks. In other words, we aim at improving the learning ability when presented with new tasks, without reducing, if not improving, the performance on old tasks. 
Following \cite{riemer_learning_2018}, the continual learning problem can be defined as the task of minimizing the loss function $L$ on the set of parameters $\theta$ for the sequence of data samples $D_t = \{(x_1,y_1),\ldots, (x_i,y_i),\ldots \}$ of the task $T_t$, where $x$ represents the input features and $y$ is the target output. For every sample pair $(x_i,y_i) \sim D_t$ and $(x_j,y_j) \sim D_t$, a positive transfer of information happens if 
$$
\nabla_\theta L(x_i,y_i) ^T \nabla_\theta L(x_j,y_j) > 0 ,
$$ 
while a negative information transfer (interference) occurs when this inner product is negative (see Fig.\ref{fig:cl_loss}). 
One way to achieve a positive learning transfer is to optimize the modified loss function 
\begin{eqnarray} \label{eq:transferloss}
\min_{\theta} &\E_{(x_i,y_i),(x_j,y_j) \sim D_1 \times D_2} \{ L(x_i,y_i) + L(x_j,y_j) \nonumber \\
&-\alpha \nabla_\theta L(x_i,y_i) ^T \nabla_\theta L(x_j,y_j) \} \  ,
\end{eqnarray} 
where in continual learning $D_1$ and $D_2$ contain the samples of tasks $t_1$ and $t_2$, respectively.

\subsection{Discriminative Bilevel Continual Learning with Episodic Memory}\label{Discriminative_Bilevel_Continual}
Training a neural network, parametrized by $\theta$, on the new task does not necessarily guarantee to retain the achieved performance on the previous tasks. A naive approach would be to preserve samples from previous tasks and use them again for the training on future tasks \cite{lopez-paz_gradient_2017}, but the re-training on a limited memory size is prone to over-fitting, despite recent effort \cite{olson2018modern}.
In order to overcome over-fitting, we split the network's parameters into two categories \textit{(i)} hyper-parameters ($\lambda$) that are shared among all tasks and \textit{(ii)} parameters ($w_t$) that are associated with the current task $T_t$, such that $\theta =(\lambda,w_t)$. 

Each new task's data $D_t$ is bisected into training $D_t^{\text{tr}}$ and validation sets $D_t^{\text{val}}$, $D_t = D_t^{\text{tr}} \bigcup D_t^{\text{val}}$.
We train $w_t$ on $D_t^{\text{tr}}$ and a relatively smaller episodic memory $M^{\text{tr}}$ sampled from the previously observed tasks' training data, similar to \cite{lopez-paz_gradient_2017}. The vector of hyper-parameters $\lambda$ is optimized on the validation data $D_t^{\text{val}}$ and a memory $M^{\text{val}}$ containing some of the previous tasks' validation samples. Since the training of $w_t$ is in the inner problem that depends on the outer problem, $w_t$ becomes dependant on the choice of $\lambda$ and will be subsequently written as $w_{\lambda,t}$. Under the assumption that both problems have the same type of loss functions $L$ for training and validation errors, we then define the discriminative Bilevel Continual Learning problem (BiCL) as
\begin{subequations}\label{bicl-discriminative} 
	\begin{eqnarray}
	\min _{\lambda} & \E_{(x,y,t),(x',y',t') \sim M^{\text{val}} \bigcup D_t^{\text{val}} } \{ \\
	& L(w_{\lambda,t},\lambda,x,y)  +L(w_{\lambda,t'},\lambda,x',y') \nonumber  \\
	&- \alpha \underbrace{\dd_\lambda L(w_{\lambda,t},\lambda,x,y) ^T   
	\dd_\lambda L(w_{\lambda,t'},\lambda,x',y')}_{\text{\normalfont inner product of hyper-gradients}} \} \nonumber  \\
	\text{s.t.} & ~ w_{\lambda,t} = \arg \min_{w_t} \E_{(x,y,t)  \sim M^{\text{tr}} \bigcup D_t^{\text{tr}} } \{ \\
	& L(w_{t},\lambda,x,y)  
	\}  . \nonumber 
	\end{eqnarray}
\end{subequations}
Since implementing Eq.\ref{bicl-discriminative} requires the computation of the gradient of the loss function on two samples and their inner product (the bracket term), we derive the solution using the reptile dynamics \cite{riemer_learning_2018} as explained in Section \nameref{sec:bilevel-reptile}. The practical implementation details are elaborated in Section \nameref{subsec:The_BiCL_algorithm}.

\subsection{Generative Bilevel Continual Learning with Episodic Memory}
\label{Generative_Bilevel_Continual}
A variational autoencoder (VAE)~\cite{kingma_auto-encoding_2013} operates with two probabilistic mappings, an encoder $X\mapsto Z$ (represented by a neural network with parameter $\phi$), and a decoder ($Z\mapsto X$ represented by another neural network with parameter $\theta$)\footnote{ These two networks will be split into hyper-parameters $\lambda$ and parameters $w_{\lambda,t}$ as in the discriminative case.}. Given a data set $\mathcal{D} = \{x^{(n)}\}_{n=1}^{N}$, an ideal VAE objective is to maximize the marginalized log-likelihood:
\begin{equation}\label{eq_ideal_VAE}
    \mathbb{E}_{p(\mathbf{x})}[\log p_\theta(\mathbf{x})].
\end{equation}
Eq.~(\ref{eq_ideal_VAE}) is, however, not tractable and is approximated by the evidence lower bound (ELBO)~\cite{kingma_auto-encoding_2013}:
\begin{equation}
    L_{\text{VAE}}(\theta, \phi) = \mathbb{E}_{q_{\phi}(z | x)} \left( 
    \log\frac{p_\theta(x|z)p(z)}{q_\phi(z|x)} \right),
\end{equation}
which can also be written as:
\begin{equation}\label{eq_VAE_obj}
\begin{split}
    L_{\text{VAE}}(\theta, \phi) &= \mathbb{E}_{q_\phi(z|x)}[\log p_\theta(x|z)]\\
    &-\mathbb{E}_{p(x)}[D_{KL}(q_\phi(z|x)\|p(z))], 
\end{split}
\end{equation}
where the first term measures the reconstruction loss, and the second one is the regularization term, which corresponds to the Kullback-Leibler (KL) divergence between the latent distribution $q_\phi(\mathbf{z}|\mathbf{x})$ and the prior distribution $p(\mathbf{z})$. Normally, we assume a fixed Gaussian prior distribution $p(\mathbf{z})$ over $\mathbf{z}$.

We define the generative BiCL problem using the VAE objective in the bilevel formulation of Eq.~\ref{eq:biproblem} and the CL objective Eq.~\ref{eq:transferloss}, as
\begin{subequations}\label{bicl-generative} 
	\begin{eqnarray}
	\min _{\lambda} & \E_{(x,t),(x',t') \sim M^{\text{val}} \bigcup D_t^{\text{val}} } \{ \\
	& L_\text{VAE}(w_{\lambda,t},\lambda,x)  +L_\text{VAE}(w_{\lambda,t'},\lambda,x') \nonumber  \\
	&- \alpha\underbrace{\dd_\lambda L_\text{VAE}(w_{\lambda,t},\lambda,x) ^T   
	\dd_\lambda L_\text{VAE}(w_{\lambda,t'},\lambda,x')}_{\text{inner product}} \}\nonumber  \\
	\text{s.t.} & ~ w_{\lambda,t} = \arg \min_{w_t} \E_{(x,t)  \sim M^{\text{tr}} \bigcup D_t^{\text{tr}} } \{ \\
	& L_{\text{VAE}}(w_{t},\lambda_{t}, x)  
	\} \ , \nonumber 
	\end{eqnarray}
\end{subequations}
where the encoder's parameters $\phi$ and decoder's parameters $\theta$ are split among $\lambda$ and $w$. How the network is split between parameters and hyper-parameters is discussed in Section ~\nameref{sec:architecture}.

\subsection{Bilevel Reptile Dynamics} \label{sec:bilevel-reptile}
We introduce a general result on how to solve the bilevel problem of Eq.\ref{eq:biproblem}, when the outer problem cost function includes the inner product of the hyper-gradient as in Eq.\ref{eq:transferloss}, which gives rise to the following transfer learning bilevel problem
\begin{subequations} \label{eq:biproblem_repltile} 
	\begin{eqnarray}
	\min _{\lambda} F(\lambda) &= \E_{s,s' \sim D^{\text{val}} \times D^{\text{val}} } \{ f(w_{\lambda,s},\lambda,s)   +f(w_{\lambda,s'},\lambda,s') \nonumber  \\
	&- \alpha_f \dd_\lambda f(w_{\lambda,s},\lambda,s)^T   \dd_\lambda f(w_{\lambda,s'},\lambda,s')\} \label{biproblem_repltile_outer}  \\
	\text{s.t.} ~ w_{\lambda} &= \arg \min_{w} \E_{b  \sim D^{\text{tr}}} \{ \label{biproblem_repltile_inner} g(w,\lambda,b) \} \ .
	\end{eqnarray}
\end{subequations}
To solve Eq.\ref{eq:biproblem_repltile} we adopt the Reptile dynamics~\cite{nichol_first-order_2018,riemer_learning_2018} while updating the parameters and hyper-parameters, using the following update rules:
\begin{subequations} \label{eq:reptilestep}
\begin{eqnarray}
\lambda_{k+1} &=& \lambda_{k}  + \beta_{\lambda} \left( p_r - \lambda_{k}  \right) \\
w_{k+1} &=& w_k + \beta_{w} \left( w_{p_r} - w_k \right) \ ,
\end{eqnarray}
\end{subequations}
with $r$ being the number of iterations between two Reptile steps, $p_{i+1} = p_{i} + \eta \dd_{\lambda} f(w_{p_i},p_i )$, $p_{0}=\lambda_{k}$, $w_{p_i}$ is the solution of the inner problem when $\lambda=p_i$ trained on the batch $D_i^{\text{tr}}$, and $\eta,\beta_{\lambda},\beta_{w}$ are learning rates. Notice that the above bilevel problem defines the general case when different objective functions are used in the outer and inner problems, $f$ and $g$. Our approach, BiCL, as presented in Eq.\ref{bicl-discriminative} and Eq.\ref{bicl-generative}  is an instantiation of Eq.\ref{eq:biproblem_repltile}.
In the following, Theorem~\ref{th:bilevelreptile} proves\footnote{Proof is in the supplementary material} how the proposed update rules do indeed lead to the solution of the general bilevel program in Eq.\ref{eq:biproblem_repltile}.

\begin{theorem}	\label{th:bilevelreptile}
	The loss function of the outer problem of Eq.\ref{eq:biproblem_repltile} can be approximated by performing the Reptile gradient step of Eq.~\ref{eq:reptilestep} on the hyper-gradient $\dd_\lambda f(w_{\lambda},\lambda)$.
\end{theorem}

\begin{algorithm}[h!]
    \caption{Bilevel Continual Learning.\\ This algorithm is one instantiation of the BiCL framework when a single head architecture is used; hence, the omission of the subscript $t$ from $w_t$.
    }
    \label{alg:BiCL}
	    
    $\text{\textbf{Continuum}}(D_t)$: this function returns training and validation batches from $D_t$\\
    $\text{\textbf{Batch-Sample}}(B,M)$: takes as input the memory $M$ and the current batch $B$ and returns $b$ sampled batches from the union $M \bigcup B$ \\
    $\text{\textbf{Reservoir}} (M ,  B^\text{tr} \bigcup B^\text{val } ) $:
    Alogrithim \ref{alg:reservoir} extends the memory $M$ with samples from $B^\text{tr} \bigcup B^\text{val }$
	\begin{algorithmic}[1]
		\Procedure{BiCL}{$\{D_1,\dots,D_{T}\}$}
		\State {$\lambda, w, M \gets \text{Init}(),\text{Init}(),\{\}$ } \Comment{Initialization of the parameters and the hyper-parameters}
		\For {$t=1 \dots T$}		
		\State $w_0',\lambda_0'    \gets w, \lambda$
		\For {$B^\text{tr},B^\text{val } \gets \text{Continuum}( D_t)$}				
		
		\State $w_0,\lambda_0    \gets w, \lambda$
		\State $B^\text{tr}_1,\dots,B^\text{tr} _b \gets \text{Batch-Sample}(B^\text{tr},M)$
		\State $B^\text{val}_1,\dots,B^\text{val}_b \gets  \text{Batch-Sample}(B^\text{val} ,M)$
		\For {i=1\dots,b}
			\For {$k=1 \to K$} \Comment{$K$ ADAM iterations}	
			\State $w_{k}  \gets \text{ADAM}(w_{k-1},\lambda,B^\text{tr}_i)$
			\EndFor	
			\State $\alpha \gets \nabla_w L(w_K,\lambda,B^\text{val}_i)$ 
			\State $p \gets \nabla_\lambda L(w_K,\lambda,B^\text{val}_i)$
			\For {$k = K  \dots  1$}
			\State $p \gets p-\eta  \nabla_{\lambda} \nabla_{w} L(w_{k-1},\lambda,B^\text{tr}_i)\alpha$
			\State $ \alpha \gets [I-\eta \nabla_{w} \nabla_w L(w_{k-1},\lambda,B^\text{tr}_i) ]\alpha $
			\EndFor
			\State $\lambda \gets \lambda + \eta p$					
		\EndFor
		\State $M \gets \text{Reservoir} (M ,  B^\text{tr} \bigcup B^\text{val } ) $ \label{mk:update}
		\State $\lambda \gets \lambda_0 + \beta_\lambda (\lambda -\lambda_0 )$ \label{mk:reptile1} \Comment{batch level reptile}
		\State $w \gets w_0 + \beta_w (w-w_0)$		
		\EndFor	
		\State $\lambda \gets \lambda_0^{\prime} + \beta_\lambda^{\prime} (\lambda -\lambda_0^{\prime} )$ \label{mk:reptile2}  \Comment{task level reptile}
		\State $w \gets w_0^{\prime} + \beta_w^{\prime} (w-w_0^{\prime})$			
		\EndFor
		\Return $\lambda, w$
		\EndProcedure
	\end{algorithmic}
\end{algorithm}

\subsection{The BiCL Algorithm}\label{subsec:The_BiCL_algorithm}
In the following, we show the practical implementation details that make our work reproducible and lead to a better understanding of the BiCL framework. 
For simplicity, we show the framework in the Continuum scenario \cite{lopez-paz_gradient_2017}. Continuum assumes that data samples arrive as triplets $(x_i,y_i,t_i,y)$, where $(x_i,y_i)$ is data point, $t_i$ is a task identifier such that $(x_i,y_i) \sim P_{t_i}(X,Y)$. In this setting, a data sample is observed only once, moreover, a switch to a new task means that samples from previous tasks will not be observed again. This, however, does not restrain the usage of a bounded size memory to keep small snapshots of previous tasks.


Algorithm~\ref{alg:BiCL} depicts the required steps of implementing a solver of Eq.\ref{bicl-discriminative}. Since we consider the Continuum scenario, tasks are sequentially observed (line 3), where each task's data $D_t$ is observed as a sequence but presented as accumulated batches using the function $\text{Continuum}( D_t)$ that returns training and validation batches (line 5).

A set of $b$ training and validation batches are created using the function $\text{Batch-Sample}(B,M)$ which takes as input the memory $M$ and the current batch (training batch $B=B^\text{tr}$ line 7, and training batch $B=B^\text{tr}$ line 8). Thereafter, the model's parameters and hyper-parameters are updated on each sampled batch (line 9-17).

Lines 10-17 compute the new hyper-parameters $\lambda$: Lines 10-11 update the parameters $w_{\lambda,t}$ (inner problem) using ADAM dynamics, and lines 12-17 compute of the hyper-gradient using the reverse hyper-gradient method \cite{franceschi_bilevel_2018}.

The algorithm employs an episodic memory $M = M^{\text{tr}} \bigcup M^{\text{val}}$ based on the reservoir sampling procedure \cite{vitter_random_1985}, defined in Alg.\ref{alg:reservoir} for completeness \footnote{Presented in the supplementary material}. The reservoir procedure mimics a uniform sampling probability on an unbounded stream of samples. Other sampling ways can be alternatively employed based on clustering methods such as K-center clustering \cite{gonzalez_clustering_1985} as proposed in \cite{nguyen_variational_2017}.

Finally, lines 19-20 and 21-22 apply the Reptile step, according to Theorem~\ref{th:bilevelreptile}, to implement the outer cost function of problem Eq.\ref{bicl-discriminative} (see Sec.\ref{sec:bilevel-reptile}).

\section{Experiments}
\label{Experiments}
In the following, we first start by discussing how neural network architecture can be used in practice to fulfill the requirement of the BiCL framework. Thereafter, we illustrate the experimental results of evaluating the proposed continual learning framework in both the discriminative and generative setting. Both types of experiments operate on sequences of tasks from various datasets.
\subsection{Practical Issues: Splitting the Network's Parameters} \label{sec:architecture}
The BiCL framework proposes to split the neural network parameters $\theta$ in two parts $\theta = (\lambda,w)$, where $\lambda$ is the vector of hyper-parameters while $w$ holds the parameters. 
\subsubsection{Discriminative Case}
\begin{figure}[h]
	\centering
	\includegraphics[width=7cm]{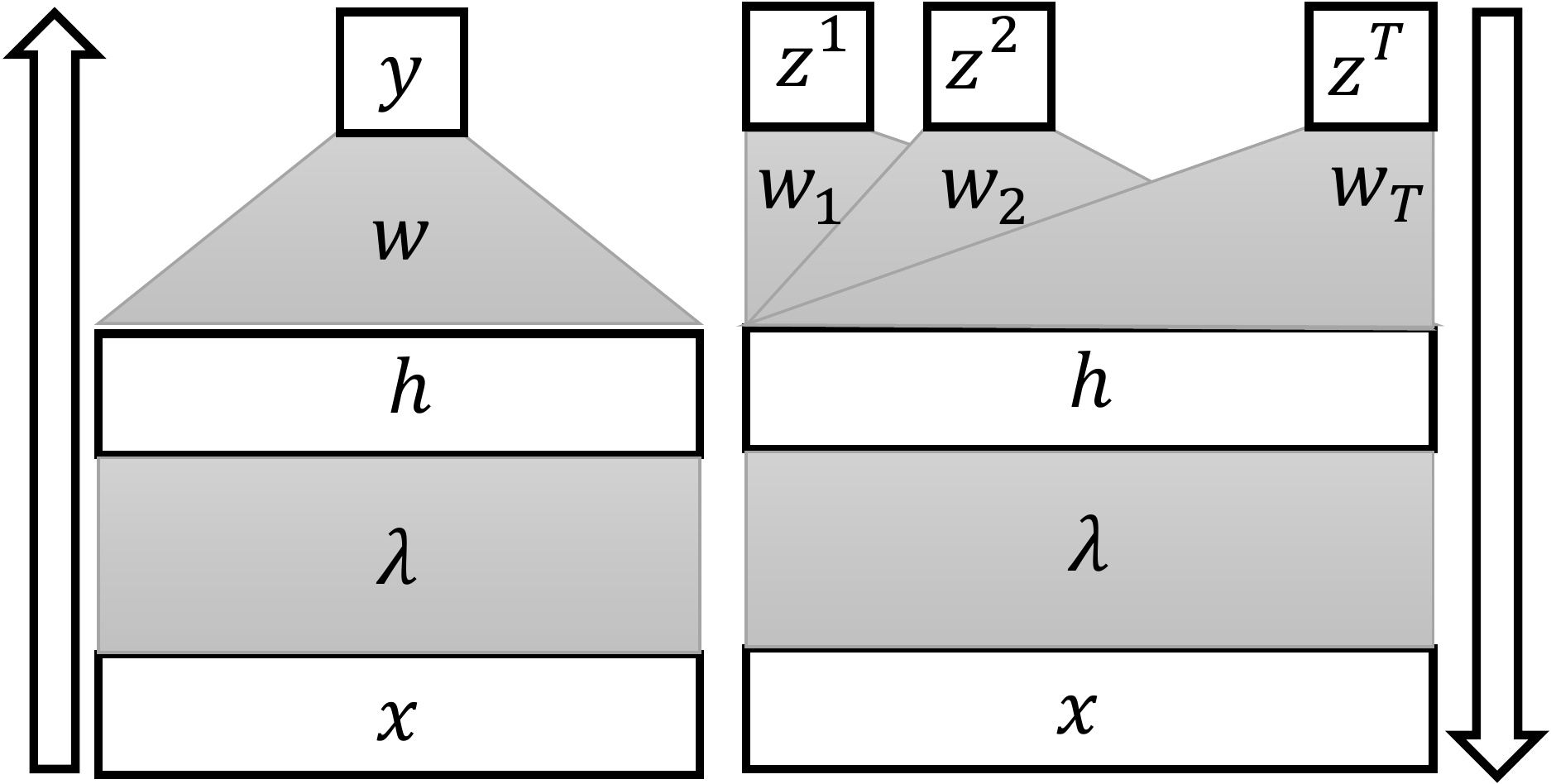}
	\caption{BiCL Continual Learning Discriminative (left) and Generative (right) Models}
	\label{fig:BiCL_figs}
\end{figure}

An intuitive way to splitting the network's parameters is to divide them horizontally. Fig.\ref{fig:BiCL_figs}(a) shows the hidden layers as hyper-parameters $\lambda$ and the single-head output layer as the  parameters $w$. Similar to \cite{nguyen_variational_2017}, the arrow indicates the direction of the forward propagation (input to output).
Alternatively, Fig~\ref{fig:BiCL_figs_reverse}(a) shows a division where the roles of $\lambda$ and $w$ are inverted.
Other splitting schemes can also be considered such as vertically dividing the network into two (or more) parallel networks (similar to ensemble networks).

\subsubsection{Generative Case}
Similar to the discriminative problem, it has to be specified, here,  which part of the network represents the parameters and which one is for the hyper-parameters.
In the generative case, we have the encoder $q_\phi(z|x)$ and decoder $p_\phi(x|z)$ networks . 
Fig.~\ref{fig:BiCL_figs}(b) shows how the encoder network can be considered as task-specific \cite{nguyen_variational_2017}, i.e., as the parameters $w$. Whereas the multi-head decoder architecture is shared among tasks, and, hence, the shared output layer becomes hyper-parameters $\lambda$. Fig~\ref{fig:BiCL_figs_reverse}(b)\footnote{In supplementary material} shows the alternative case when the roles of the decoder and encoder are exchanged.

\begin{table*}[h]
	\centering
	\caption{Results in terms of LA, RA and BTI on the MNIST rotations and permutations, Fashion-MNIST permutations and the NotMNIST permutations datasets over 10 tasks with different with various budget (\cite{riemer_learning_2018}). RO = Rotation Online, PO=Permutation Online}
\begin{tabularx}{\textwidth}{X X | X X X |X X X |X X X |X X X }

\toprule	
    \multicolumn{5}{c}  \text{MNIST RO}
    & \multicolumn{3}{|c}  \text{MNIST PO} 
    & \multicolumn{3}{|c}  \text{Fashion-MNIST PO}
    & \multicolumn{3}{|c}  \text{NotMNIST PO } \\
    \midrule
    & \text{Size} 
    &  \text{RA} &  \text{LA} &  \text{BTI} 
    &  \text{RA} &  \text{LA} &  \text{BTI} 
    &  \text{RA} &  \text{LA} &  \text{BTI} 
    &   \text{RA} &  \text{LA} &  \text{BTI} \\
    \midrule
    
\text{Online}  & & { 58.55}   \scriptsize{(1.71)}    &{ 87.15}   \scriptsize{(0.5)}   &{ 28.61}   \scriptsize{(1.73)}    &{  58.39}  \scriptsize{(1.24)}    &{ 83.57}   \scriptsize{(0.44)}    &{25.18}    \scriptsize{(1.59)} & {  43.37}   \scriptsize{(3.58)}    &{ 73.77}   \scriptsize{(0.88)}   &{ 30.4}   \scriptsize{(3.94)}    &{ 38.81}  \scriptsize{(5.2)}    &{ 80.37}   \scriptsize{(1.05)}    &{41.56}    \scriptsize{(4.75)}  \\
\midrule
\text{Independent}  & & { 83.45}   \scriptsize{(0.46)}    &{ 83.45}   \scriptsize{(0.46)}   &{ 0}   \scriptsize{(0)}    &{  83.75}  \scriptsize{(0.61)}    &{ 83.75}   \scriptsize{(0.61)}    &{0}    \scriptsize{(0)} & {  73.97}   \scriptsize{(0.55)}    &{ 73.97}   \scriptsize{(0.55)}   &{ 0}   \scriptsize{(0)}    &{ 81.57}  \scriptsize{(0.6)}    &{ 81.57}   \scriptsize{(0.6)}    &{0}    \scriptsize{(0)}  \\
\midrule
\text{EWC}  & & { 67.53}   \scriptsize{(0.8)}    &{ 72.99}   \scriptsize{(0.68)}   &{ 5.46}   \scriptsize{(1.15)}    &{  71.64}  \scriptsize{(1.09)}    &{ 80.46}   \scriptsize{(0.19)}    &{8.82}    \scriptsize{(1.09)} & {  11.47}   \scriptsize{(1.23)}    &{ 13.7}   \scriptsize{(2.22)}   &{ 2.23}   \scriptsize{(2.3)}    &{ 68.14}  \scriptsize{(3.43)}    &{ 80.62}   \scriptsize{(1.36)}    &{12.48}    \scriptsize{(2.08)}  \\
\midrule
\text{GEM}  & 5120& { 87.24}   \scriptsize{(1.56)}    &{ 86.08}   \scriptsize{(0.29)}   &{ 3.18}   \scriptsize{(1.30)}    &{  83.91}  \scriptsize{(0.33)}    &{ 80.74}   \scriptsize{(0.28)}    &{-3.18}    \scriptsize{(0.53)} & {  74.29}   \scriptsize{(0.57)}    &{ 68.55}   \scriptsize{(0.79)}   &{ -5.74}   \scriptsize{(0.64)}    &{ 82.61}  \scriptsize{(0.57)}    &{ 79.17}   \scriptsize{(0.5)}    &{-3.44}    \scriptsize{(0.76)}  \\
\text{}  & 500& { 79.88}   \scriptsize{(1.03)}    &{ 85.65}   \scriptsize{(0.31)}   &{ 5.77}   \scriptsize{(1.17)}    &{  73.63}  \scriptsize{(0.61)}    &{ 80.38}   \scriptsize{(0.45)}    &{6.75}    \scriptsize{(0.83)} & {  65.6}   \scriptsize{(1.41)}    &{ 67.73}   \scriptsize{(0.57)}   &{ 2.12}   \scriptsize{(1.21)}    &{ 73.59}  \scriptsize{(1.04)}    &{ 80.96}   \scriptsize{(0.86)}    &{7.37}    \scriptsize{(0.62)}  \\
\text{}  & 200& { 74.4}   \scriptsize{(1.32)}    &{ 84.91}   \scriptsize{(0.3)}   &{ 10.51}   \scriptsize{(1.12)}    &{  61.21}  \scriptsize{(3.05)}    &{ 80.04}   \scriptsize{(0.18)}    &{18.83}    \scriptsize{(2.95)} & {  56.26}   \scriptsize{(1.68)}    &{ 67.52}   \scriptsize{(0.97)}   &{ 11.26}   \scriptsize{(2.17)}    &{ 67.57}  \scriptsize{(2.1)}    &{ 80.2}   \scriptsize{(1.05)}    &{12.63}    \scriptsize{(2.21)}  \\
\midrule
\text{MER}  & 5120& { \bf 92.16}   \scriptsize{(0.09)}    &{ \bf 90.31}   \scriptsize{(0.08)}   &{\bf -1.84}   \scriptsize{(0.09)}    &{  \bf 88.81}  \scriptsize{(0.09)}    &{ \bf 88.92}   \scriptsize{(0.12)}    &{0.11}    \scriptsize{(0.14)} & {  77.08}   \scriptsize{(0.30)}    &{ 74.87}   \scriptsize{(0.13)}   &{ \bf -2.21}   \scriptsize{(0.25)}    &{ 85.9}  \scriptsize{(0.20)}    &{ 85.31}   \scriptsize{(0.14)}    &{ \bf -0.59}    \scriptsize{(0.12)}  \\
\text{}  & 500& { \bf 87.85}   \scriptsize{(0.25)}    &{ 87.56}   \scriptsize{(0.18)}   &{ \bf -0.29}   \scriptsize{(0.32)}    &{  \bf 82.73}  \scriptsize{(0.38)}    &{ 85.18}   \scriptsize{(0.37)}    &{\bf 2.45}    \scriptsize{(0.5)} & {  71.52}   \scriptsize{(1.11)}    &{ 74.38}   \scriptsize{(0.43)}   &{ 2.86}   \scriptsize{(1.07)}    &{ \bf 80.02}  \scriptsize{(0.47)}    &{ 83.25}   \scriptsize{(0.13)}    &{\bf 3.23}    \scriptsize{(0.46)}  \\
\text{}  & 200& { 84.68}   \scriptsize{(0.35)}    &{ 84.66}   \scriptsize{(0.40)}   &{ -0.02}   \scriptsize{(0.43)}    &{  79.23}  \scriptsize{(0.64)}    &{ 84.44}   \scriptsize{(0.46)}    &{5.21}    \scriptsize{(0.63)} & {  66.99}   \scriptsize{(1.11)}    &{ 72.74}   \scriptsize{(0.90)}   &{ 5.75}   \scriptsize{(0.64)}    &{ 76.61}  \scriptsize{(0.98)}    &{ 83.26}   \scriptsize{(0.36)}    &{6.65}    \scriptsize{(0.97)}  \\
\midrule
\text{BiCL}  & 5120& { 91.02}   \scriptsize{(0.16)}    &{\bf 90.16}   \scriptsize{(0.12)}   &{ -0.86}   \scriptsize{(0.16)}    &{  87.57}  \scriptsize{(0.19)}    &{ 88.09}   \scriptsize{(0.27)}    &{\bf -0.52}    \scriptsize{(0.2)} & {  \bf 77.7}   \scriptsize{(0.28)}    &{\bf 77.62}   \scriptsize{(0.16)}   &{ 0.08}   \scriptsize{(0.28)}    &{ \bf 87.15}  \scriptsize{(0.17)}    &{ \bf 87.45}   \scriptsize{(0.07)}    &{0.30}    \scriptsize{(0.10)}  \\
\text{}  & 500& { 86.88}   \scriptsize{(0.12)}    &{ \bf 88.53}   \scriptsize{(0.14)}   &{ 1.65}   \scriptsize{(0.11)}    &{  82.42}  \scriptsize{(0.24)}    &{ \bf 85.85}   \scriptsize{(0.05)}    &{3.43}    \scriptsize{(0.28)} & {  \bf 72.81}   \scriptsize{(0.31)}    &{ \bf 74.57}   \scriptsize{(0.36)}   &{ \bf 1.76}   \scriptsize{(0.42)}    &{ \bf 80.09}  \scriptsize{(0.49)}    &{ \bf 83.76}   \scriptsize{(0.22)}    &{3.66}    \scriptsize{(0.7)}  \\
\text{}  & 200& { 82.96}   \scriptsize{(0.10)}    &{ \bf 89.18}   \scriptsize{(0.08)}   &{ 6.22}   \scriptsize{(0.06)}    &{  \bf 80.25}  \scriptsize{(0.45)}    &{ 83.71}   \scriptsize{(0.24)}    &{\bf 3.46}    \scriptsize{(0.62)} & {  \bf 68.26}   \scriptsize{(0.36)}    &{ \bf 73.65}   \scriptsize{(0.24)}   &{ \bf 5.39}   \scriptsize{(0.37)}    &{ \bf 77.52}  \scriptsize{(0.11)}    &{\bf  83.23}   \scriptsize{(0.20)}    &{\bf 5.71}    \scriptsize{(0.15)}  \\
    
    \bottomrule
\end{tabularx}
    \label{tb:comparison}
\end{table*}

\begin{figure*}[h]
	\centering
	\includegraphics[width= \textwidth]{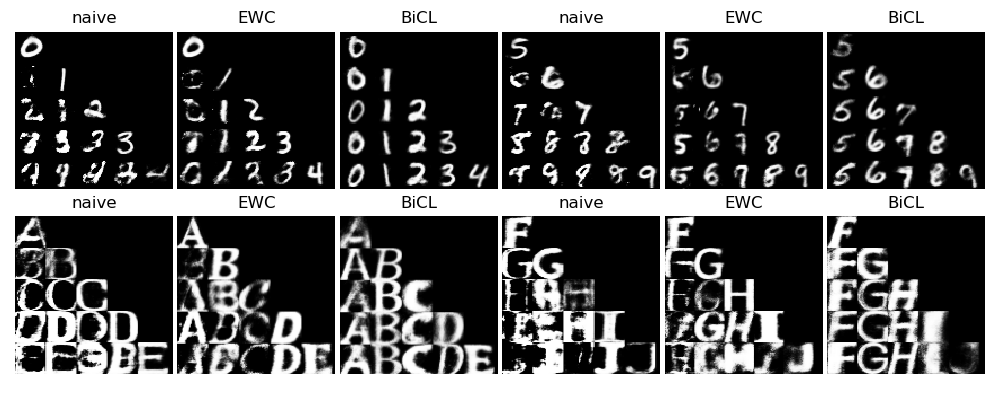}
	\caption{Generated images from each generator after training after each task. The first row of images shows the results on the MNIST data, and the second row is for the notMNIST data.
	In each image, each cell (row $\times$ column) shows a generated letter (or character) by a generator after being trained on all classes on the same row. Characters and numbers generated by BiCL are clearer than those generated by EWC. See for example the numbers $0$ and $5$ for MNIST, and the letters \textit{A} and \textit{F} for notMNIST.
	}
	\label{fig:BiCL_figs_generative_visualization}
\end{figure*}





\subsection{Discriminative Online Continual Learning}
In this type of experiment, we adopt the online setup used in \cite{riemer_learning_2018}, where tasks are observed continuously and each method is permitted to observe every data sample only once.
We evaluate the performance of our method when the memory buffer takes the size of $200$, $500$, and $5120$ against two baselines and three state-of-the-art methods. Four experiments were performed; each involves a different dataset with ten tasks while limiting the number of training examples per task to 1000 samples.
To compare with recent approaches, we adopt the continuum concept (one realization of the online setup) as defined in \cite{lopez-paz_gradient_2017}, which allows the input data to be buffered for the current task only.

\subsubsection{Discriminative Baselines}
We evaluate the performance of BiCL against that of two baselines and three competitive methods used in \cite{riemer_learning_2018}. 
These include: {\bf Online}: a method that simply applies stochastic gradient descendent (SGD) on each sample continuously. {\bf Independent}: this method trains independent network, one for each task, with reduced number of neurons. {\bf EWC} Elastic Weight Consolidation \cite{kirkpatrick_overcoming_2016}, {\bf GEM} Gradient Episodic Memory \cite{lopez-paz_gradient_2017}, and {\bf MER} Meta Experience Reply which were explained in Section \nameref{related_work}.
\subsubsection{Discriminative Neural Network Architecture}
As in \cite{lopez-paz_gradient_2017}, 
we use a single head fully connected neural network architecture with two hidden layers of size $100$, $28 \times 28$ inputs, and $10$ outputs. The hidden layers use ReLu activation function. 
For the categorical discriminative case in Eq.\ref{bicl-discriminative}, the outer problem minimizes the softmax cross-entropy on the validation data\footnote{More specifically, it minimizes the maximum softmax cross-entropy over all samples in the validation set}. The inner problem minimizes the average softmax cross-entropy across training samples.

\subsubsection{Datasets}
The evaluation is performed on {\bf  MNIST Permutations} dataset, which is a variation of MNIST proposed by \cite{kirkpatrick_overcoming_2016}, where each task contains a fixed permutation of the MNIST's input pixels. {\bf MNIST Rotations} \cite{lopez-paz_gradient_2017} is another continual learning variant of MNIST where for each task the MNIST images are rotated by a fixed angle between $0$ and $180$ degrees. 
{\bf Fashion-MNIST} \cite{xiao2017_online} and {\bf notMNIST} \cite{bulatov_machine_2011} datasets share the same format of MNIST but contain images of Zalando's clothing products and letters, respectively.
\subsubsection{Discriminative Metrics}
The performance of the CL methods is measured trough {\bf Learning Accuracy (LA)} that is the average accuracy on each tasks' test data directly after learning that task. {\bf Retained Accuracy (RA)} is the average accuracy on all tasks, after the training on the last task. 
{\bf Backward Transfer of Information (BTI)} is the difference between the learning accuracy and the retained accuracy. More formally, LA and RA are defined as follows \cite{chaudhry_efficient_2019}:

\begin{alignat}{2}
LA  = \frac{1}{T} \sum_{i=1}^{T} a_{i,i} \ \ , RA = \frac{1}{T} \sum_{i=1}^{T} a_{T,i} \ ,
\label{eq:LA_RA}
\end{alignat}
where $a_{j,i}$ is the accuracy on the $i$th task after train on the $j$th task.

\subsubsection{Discriminative Results and Analysis}
Table~\ref{tb:comparison} shows the results of comparing the performance of BiCL performance with that of the state-of-the-art CL methods on various datasets in terms of RA, LA, and BTI. Different memory sizes (when applicable) are used 200, 500, and 5120 instances. The best achieved result across each setting is marked in bold. Multiple winners are marked if they do not differ statically.

On the first two data sets, permutation and rotation MNIST, we notice that BiCL often achieves a better LA while having a comparable RA performance with that of MER. On Fashion-MNIST, BiCL shows superiority in both LA and RA in all different settings of memory sizes. A similar observation can be observed on the NotMNIST dataset.

\subsection{Generative Continual Learning}
We evaluate the generative models in a continual learning scenario on two datasets: MNIST for digit generation and notMNIST for character generation. Since notMNIST does not have a predefined splitting between the train and test datasets and has fewer samples, we reduce MNIST and notMNIST to 2000 samples for each class and we split the train and test datasets with ratio 0.9. We then evaluate the ability for overcoming catastrophic forgetting on 5 tasks separately (i.e. 5 digits and 5 characters respectively). Here each class is presented as a task.

We adopt the network architecture of  \cite{nguyen_variational_2017}. The model consists of shared generator components as well as task-specific heads which includes a $4$ layers encoder and a generator head. Each layer is a fully-connected layer with $500$ hidden units and the dimensionality for the latent vector is $50$. We use the proposed BiCL framework to train the model, where we set the weights of the head layers as parameters in the inner problem and the weights of shared layers as hyper-parameters in the outer problem. Besides, we use a sample memory of size $500$ in total. 

We compare our BiCL with the naive online learning using the standard VAE objective, and with EWC after setting its hyper-parameter to $\lambda=10$ as in \cite{nguyen_variational_2017}. Samples from the generative models attained at different time steps are shown in Fig.~\ref{fig:BiCL_figs_generative_visualization}. 



EWC achieves slightly better log-likelihood for the current task (see supplementary material), but BiCL has a superior long-term retain performance on previous tasks in both MNIST and notMNIST, which indicates improved capacity of reducing catastrophic forgetting. In Fig.~\ref{fig:BiCL_figs_generative_visualization}, BiCL produces high-quality results, whereas EWC fails in some tasks (letter or number, e.g. last line, first column). The qualitative and numerical evaluations all confirm that BiCL is also suitable for generative setting and outperforms the baseline models.

\section{Final Remarks and Conclusion}
\label{ Conclusions}

Bilevel has been proposed for multi-task learning \cite{flamary2014learning,franceschi_bilevel_2018,frecon2018bilevel} and deep learning \cite{jenni2018deep}. In this paper, we propose the use of bilevel optimization of the continual learning problem for two reasons. Firstly, bilevel formulation differentiates the training of task-specific (inner problem) and task agnostic (outer-problem) parts of any neural network architecture. Secondly, since in continual learning data arrive sequentially, bilevel jointly and sequentially computes the hyper-gradient on the validation and training data while fitting the parameters on the training data. Otherwise, conventional hyperparameter optimization requires multiple passes over the validation and training datasets. 


The paper shows how the bilevel problem formulation allows transfer-learning among tasks in deep neural networks for both discriminative and generative learning problems, where the outer problem guides the update of the inner problem such that the network retains knowledge of previous tasks.

Experimental results show that BiCL provides superior performance in terms of retained accuracy in 75\% of all cases when the memory size is very constrained while offering competitive learning accuracy. Moreover, BiCL generates higher quality images across all tasks qualitatively and quantitatively than baseline methodologies. Hence, BiCL shows a robust solution towards reducing the effect of catastrophic forgetting. 
The general solution framework used in BiCL could be applied in other ML problems, as meta-learning, where tasks are available at the same time and a new model needs to be generated for a specific dataset. Further, different architectures (i.e. split of parameters/hyper-parameters) could be used and experimented using the outlined framework.





\bibliographystyle{abbrv}
\bibliography{bcl}

\appendix
\section{Bilevel Continual Learning \\Supplementary material }
\subsection{Proofs}
\begin{proof} [Theorem 1.]
 We follow \cite{riemer_learning_2018} and expand the total derivative of Eq.\ref{biproblem_repltile_outer} with respect to $\lambda$. 
 
 The basic shape of the hyper-gradient is 
\begin{equation}
\dd _{\lambda} f(w_\lambda,\lambda) = \nabla_\lambda f(w_\lambda,\lambda)+  \nabla_\lambda w_\lambda \nabla_w f(w_\lambda,\lambda)
\end{equation}
where $\dd$ is the total derivative and $\nabla$ the partial. 
We define $g(\lambda) = f(w_\lambda,\lambda)$ and apply two steps of SGD on this function
\begin{eqnarray*}
\lambda_1 &=& \lambda_0 - \alpha \dd _{\lambda} g_0(\lambda_0) \\
\lambda_2 &=& \lambda_1 - \alpha \dd _{\lambda} g_1(\lambda_1) 
\end{eqnarray*}
where $g_i(\lambda)$ is the gradient of the hyper-parameter evaluated on the $i$-batch. We now use the first order Taylor expansion with respect to $\lambda$ 
\begin{eqnarray*}
\dd _{\lambda} g_1(\lambda_1)  &\approx & \dd _{\lambda} g_1(\lambda_0) + \dd^2 _{\lambda} g_1(\lambda_0)  (\lambda_1-\lambda_0) + O(\alpha^2) \nonumber \\
&\approx & \dd _{\lambda} g_1(\lambda_0) - \alpha \dd^2 _{\lambda} g_1(\lambda_0)   \dd _{\lambda} g_0(\lambda_0) +O(\alpha^2) 
\end{eqnarray*}
and finally compute the reptile step
\begin{alignat*}{2}
p_\text{reptile} &=& \frac1{\alpha} { (\lambda_0 - \lambda_2)}  = \dd _{\lambda} g_0(\lambda_0) + \dd _{\lambda} g_1(\lambda_1) \\
&\approx & \dd _{\lambda} g_0(\lambda_0) + \dd _{\lambda} g_1(\lambda_0) - \alpha \dd^2 _{\lambda} g_1(\lambda_0)   \dd _{\lambda} g_0(\lambda_0)  
\end{alignat*}
Similar to \cite{nichol_first-order_2018}, this is equivalent to minimize in the outer objective Eq.\ref{bicl-discriminative}, when $g_i(\lambda) = E_{b \sim B_i} L(w_\lambda,\lambda,b)$ and $b = (x,y,t)$ (or $b = (x,t)$ for the generative model), since 
\begin{alignat*}{2}
 \E \{p_\text{reptile} \} &=& \E \{\dd _{\lambda} g_0(\lambda_0)\} + \E \{\dd _{\lambda} g_1(\lambda_0)\} \nonumber  \\
&& - \alpha \E \{\dd^2 _{\lambda} g_1(\lambda_0)   \dd _{\lambda} g_0(\lambda_0) \} \nonumber \nonumber  \\
&=& \E \{\dd _{\lambda} g_0(\lambda_0)\} + \E \{\dd _{\lambda} g_1(\lambda_0)\} \nonumber \nonumber  \\
&& - \alpha \E \{\dd_{\lambda} \left( \dd_{\lambda} g_1(\lambda_0) ^T  \dd _{\lambda} g_0(\lambda_0) \right)  \}
\end{alignat*}
where the last line is the gradient in the direction of the inner product of the two directions. 
\end{proof}

\subsection{Architecture}

\begin{figure}[h]
	\centering
	\includegraphics[width=7cm]{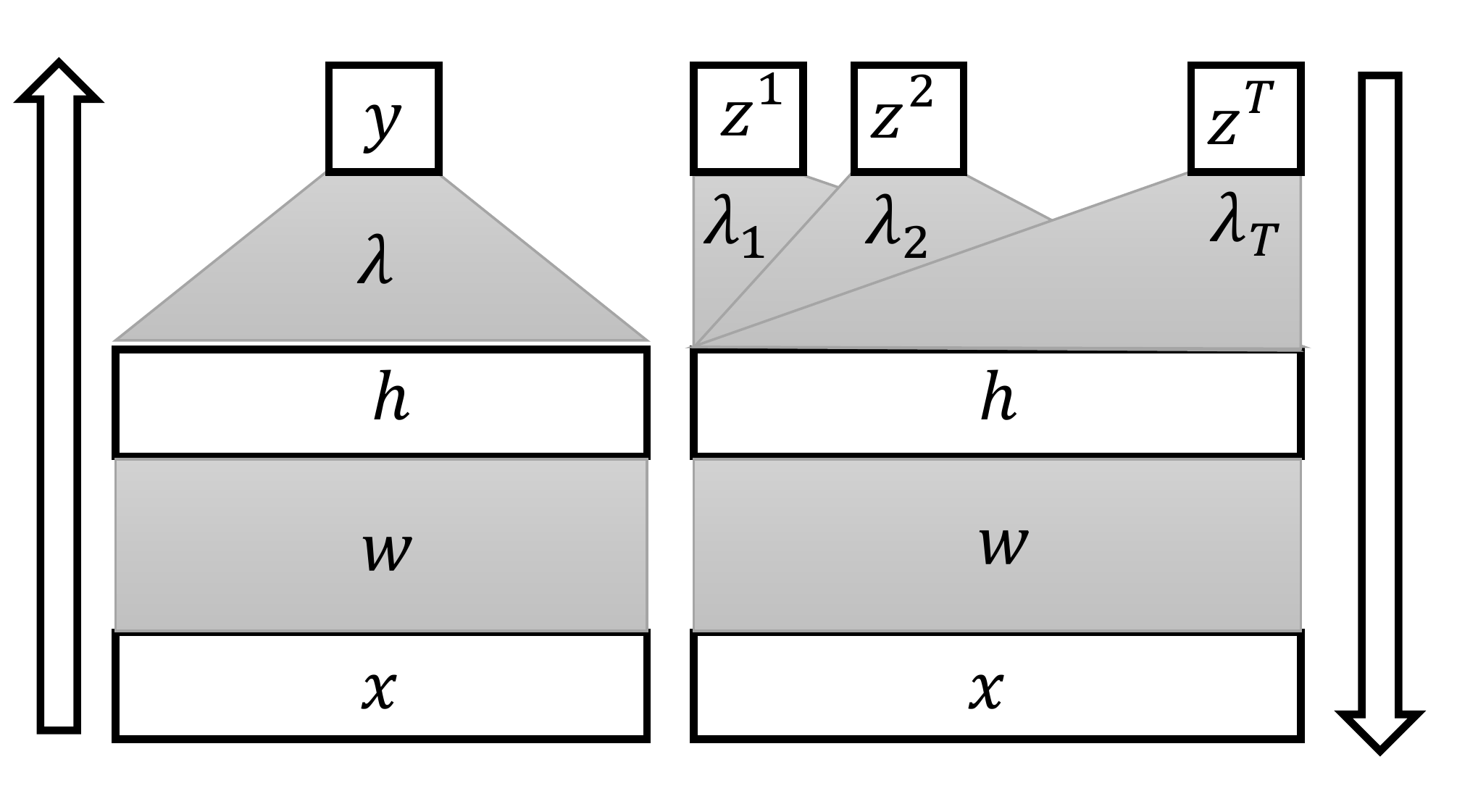}
	\caption{BiCL Continual Learning Discriminative (left) and Generative (right) Models, where hyper-parameters and parameters roles are interchanged.}
	\label{fig:BiCL_figs_reverse}
\end{figure}

\subsection{Algorithms}
In Alg.\ref{alg:reservoir}, the Reservoir Algorithm is described for completeness. 
\begin{algorithm}[h!]
	\caption{Reservoir Episodic Memory Update of maximum size of $N_\text{max}$}
	\label{alg:reservoir}
	\begin{algorithmic}[1]
		\Procedure{Reservoir}{$M,B$}
		\For{ $(x,y) \in B $}
		\State $i \gets \text{RandInt}(|M|)$
		\If{$i \le |M| \lor |M| \le N_\text{max}$}
		\State $M[i] \gets (x,y)$
		\EndIf
		\EndFor
		\Return $M$
		\EndProcedure
	\end{algorithmic}
\end{algorithm}

\subsection{Additional on Generative experiments}
\begin{figure*}[h]
	\centering
	\includegraphics[width= 0.9\textwidth]{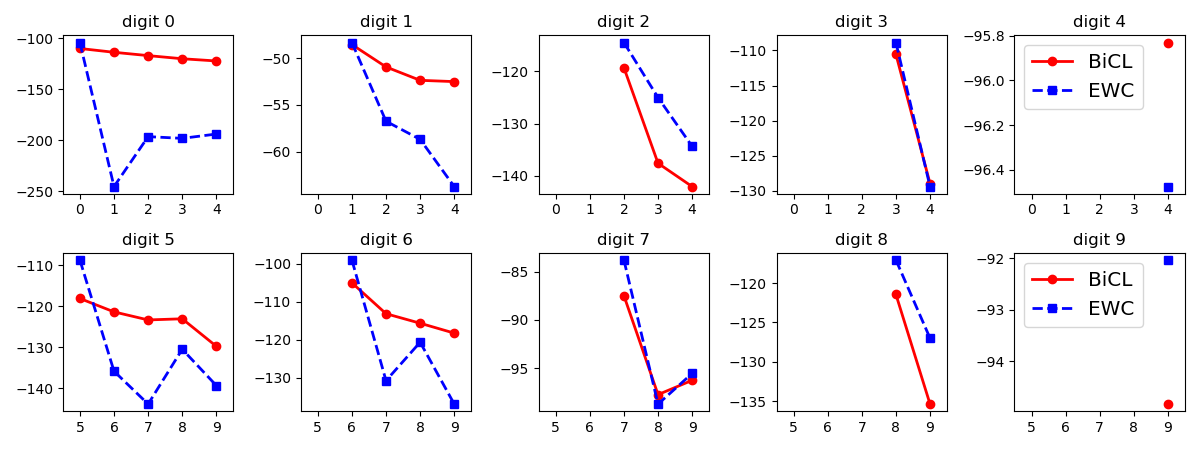}
	\caption{Test-LL results on the subset of MNIST. The higher the better.}
	\label{fig:BiCL_figs_generative_mnist}
\end{figure*}	

\begin{figure*}[h]
	\centering
	\includegraphics[width= 0.9\textwidth]{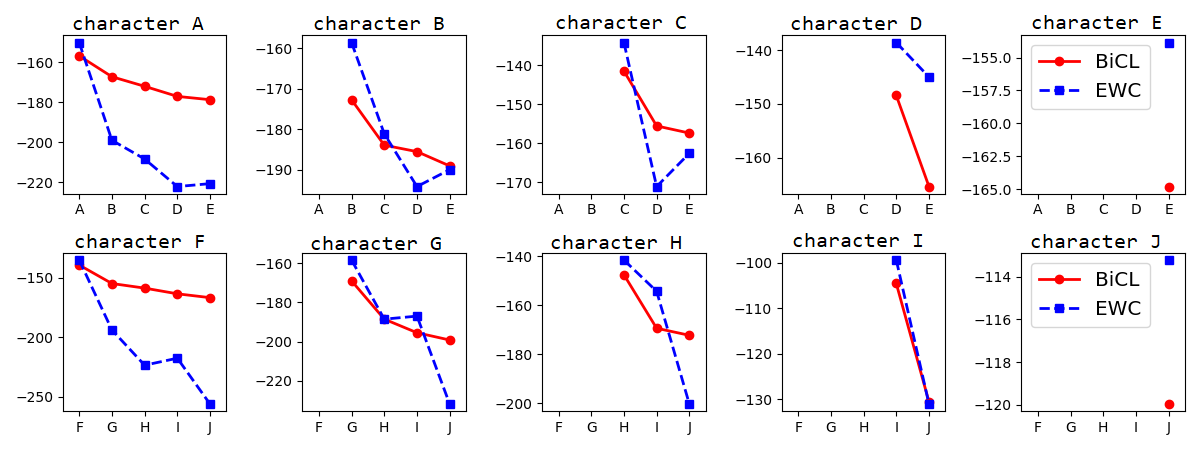}
	\caption{Test-LL results on notMNIST. The higher the better.}
	\label{fig:BiCL_figs_generative_notmnist}
\end{figure*}

Here, an importance sampling estimate of the test log-likelihood (test-LL) using $5,000$ samples is used to evaluate the model performance quantitatively. The quantitative comparison is shown in  Fig.~\ref{fig:BiCL_figs_generative_mnist} and Fig.~\ref{fig:BiCL_figs_generative_notmnist}. 

From Fig.\ref{fig:BiCL_figs_generative_mnist} and Fig.\ref{fig:BiCL_figs_generative_notmnist}, it is easy to observe that EWC achieve slightly better log-LL for the current task, but BiCL has a superior long-term retain performance on previous tasks in both MNIST and notMNIST, which indicates improved capacity of reducing catastrophic forgetting. In Fig.~\ref{fig:BiCL_figs_generative_visualization}, BiCL produces high-quality results, whereas EWC fails in some tasks (letter or number, e.g. last line, first column). The qualitative and numerical evaluations all confirm that BiCL is also suitable for generative setting and outperforms the baseline models.

\end{document}